\documentclass[runningheads]{llncs}
\usepackage[T1]{fontenc}

\usepackage{graphicx}
\usepackage{amsmath, amssymb}

\begin{document}

\title{Similarity-Based Self-Construct Graph Model for Predicting Patient Criticalness Using Graph Neural Networks and EHR Data}
\author{Mukesh Kumar Sahu\inst{1}\orcidID{0009-0005-4007-7648}
\and Pinki Roy \inst{2}}
\authorrunning{Mukesh Kumar Sahu}
\institute{National Institute of Technology Silchar, Assam, India\\
\email{mukesh\_pg23@cse.nits.ac.in} and  
\email{pinki@cse.nits.ac.in}}

\maketitle

\begin{abstract}
Accurately predicting the criticalness of ICU patients (such as in-ICU mortality risk) is vital for early intervention in critical care. However, conventional models often treat each patient in isolation and struggle to exploit the relational structure in Electronic Health Records (EHR). We propose a \textbf{Similarity-Based Self-Construct Graph Model (SBSCGM)} that dynamically builds a patient similarity graph from multi-modal EHR data, and a \textbf{HybridGraphMedGNN} architecture that operates on this graph to predict patient mortality and a continuous criticalness score. SBSCGM uses a hybrid similarity measure (combining feature-based and structural similarities) to connect patients with analogous clinical profiles in real-time. The HybridGraphMedGNN integrates Graph Convolutional Network (GCN), GraphSAGE, and Graph Attention Network (GAT) layers to learn robust patient representations, leveraging both local and global graph patterns. In experiments on 6,000 ICU stays from the MIMIC-III dataset, our model achieves state-of-the-art performance (AUC-ROC $0.94$) outperforming baseline classifiers and single-type GNN models. We also demonstrate improved precision/recall and show that the attention mechanism provides interpretable insights into model predictions. Our framework offers a scalable and interpretable solution for critical care risk prediction, with potential to support clinicians in real-world ICU deployment.
\end{abstract}

\section{Introduction}

The widespread availability of large-scale electronic health records (EHRs) has created new opportunities for predictive modeling in critical care. However, ICU data is inherently high-dimensional, heterogeneous, and temporally dynamic, posing significant challenges for conventional learning systems. Most predictive models including logistic regression and Transformer-based EHR models (e.g., Med-BERT, Hi-BEHRT) treat patients independently and neglect underlying similarities between clinical trajectories, thereby missing relational patterns indicative of deterioration.

To address this, we propose modeling patients as a dynamic graph, where edges represent clinical similarity. We introduce the \textbf{Similarity-Based Self-Construct Graph Model (SBSCGM)}, which builds a patient similarity graph in real time using a hybrid similarity function, and present \textbf{HybridGraphMedGNN}, a novel GNN architecture that integrates GCN, GraphSAGE, and GAT layers to exploit both local and global graph structures for ICU outcome prediction.

Our main contributions are as follows. First, we propose a dynamic, data-driven patient graph construction strategy that evolves with new ICU data, offering adaptability beyond static graph models. Second, we develop a hybrid similarity measure that combines cosine-based feature similarity and Jaccard-based structural similarity, allowing for robust edge formation. Third, we design a multi-architecture GNN that fuses the strengths of GCN, GraphSAGE, and GAT to generate interpretable, multi-scale embeddings. Finally, through multi-task training (mortality classification and severity regression), our method achieves state-of-the-art performance on the MIMIC-III dataset, outperforming classical models and single-type GNNs. Ablation studies further show the advantage of integrating static and temporal features in the graph.

By linking patients with analogous clinical profiles and leveraging GNN-based reasoning, this work advances explainable, high-fidelity risk prediction for ICU patients and lays the groundwork for real-time, graph-based decision support in critical care.

\section{Related Work}

We review foundational advancements in four key domains relevant to our work: (1) Graph Neural Networks (GNNs) for ICU risk modeling, (2) dynamic graph construction in clinical settings, (3) multimodal integration of EHR data, and (4) explainability in graph-based healthcare AI.

\subsection{GNNs for ICU Outcome Prediction}

GNNs have increasingly been adopted in critical care research for their ability to model inter-patient dependencies and uncover latent relationships across cohorts. Ma~\emph{et al.}~\cite{Ma2023} introduced a dynamic GAT-based model for ICU mortality prediction, achieving up to 1.8\% AUC improvement over static graph baselines. Boll~\emph{et al.}~\cite{Boll2024} used patient similarity graphs for heart failure prediction, while Defilippo~\emph{et al.}~\cite{Defilippo2024} demonstrated GNN utility in automating emergency triage with interpretable outputs. A systematic review by Gao~\emph{et al.}~\cite{Gao2023} consolidates these trends, highlighting GNNs’ edge over traditional models in capturing relational structure. Recent studies have further explored advanced variants such as hypergraphs~\cite{Huang2022} and early-warning systems~\cite{Zhao2023}, validating GNN robustness in complex ICU environments.

\subsection{Dynamic Graph Construction in Healthcare}

Traditional graph-based models often rely on static similarity derived from shared diagnoses or demographics, limiting their responsiveness to clinical progression. To address this, Xu~\emph{et al.}~\cite{Xu2023} proposed a temporal GNN that dynamically updates graph structure based on evolving EHR signals. Our SBSCGM framework builds on this idea, constructing a hybrid patient graph using both static features and real-time vitals. Unlike fixed-topology methods, our approach adaptively redefines connectivity to reflect the most recent patient trajectories, significantly enhancing predictive accuracy as shown in Section~\ref{sec:results}.

\subsection{Multimodal EHR Fusion}

ICU patient data is inherently heterogeneous spanning structured variables (e.g., vitals, diagnoses), semi-structured codes, and unstructured text~\cite{Johnson2016}. Zhou~\emph{et al.}~\cite{Zhou2022} proposed PM$^2$F$^2$N, which fuses clinical notes and time-series vitals through co-attention and graph-based correlation modeling. Graph representations are particularly suited for such multimodal fusion, as they allow flexible encoding of various data types within nodes and edges. While our current implementation integrates structured inputs into node features, future extensions may leverage pretrained language models such as BioBERT~\cite{Lee2020} and Med-BERT~\cite{Liu2021} for textual enrichment, or combine image features using vision-language embeddings~\cite{Rajkomar2018,Shickel2018}.

\subsection{Explainability in Clinical GNNs}

Interpretability is vital for clinician trust and regulatory acceptance. GATs~\cite{Velickovic2018} offer inherent transparency via attention weights, which quantify the influence of neighboring nodes during prediction. RETAIN~\cite{Choi2016} demonstrates how attention can uncover temporal salience in medical histories, while SHAP and other feature attribution tools~\cite{Shickel2018} are often used post hoc. In HybridGraphMedGNN, we utilize GAT-derived attention to trace peer influence in mortality scoring. Beyond metrics, we conduct error analysis on false positives and negatives to validate alignment with clinical reasoning. Future directions may include integrating GNNExplainer or counterfactual reasoning frameworks to further enhance decision interpretability in safety-critical ICU applications.

\section{Methodology}

Our proposed framework integrates two key modules: (1) the \textbf{Similarity-Based Self-Constructing Graph Model (SBSCGM)} for dynamic patient graph construction based on EHR-derived similarity metrics, and (2) the \textbf{HybridGraphMedGNN}, a heterogeneous graph neural network designed to perform both mortality classification and severity regression on the constructed graph.

\begin{figure}[tb]
    \centering
    \includegraphics[width=\textwidth]{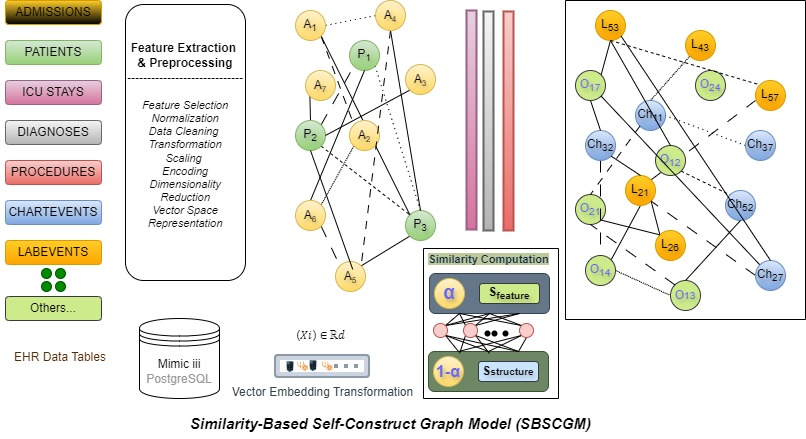}
    \caption{Overview of the SBSCGM framework. Structured EHR data are transformed into feature vectors, pairwise hybrid similarities are computed, and high-similarity links form a dynamic patient graph used as input to the GNN for clinical outcome prediction.}
    \label{fig:sbscgm_overview}
\end{figure}

\subsection{Patient Similarity Graph Construction (SBSCGM)}

We define the ICU cohort as an undirected, weighted graph $G = (V, E)$, where each node $v \in V$ represents a patient, and each edge $(u, v) \in E$ encodes the clinical similarity between patients $u$ and $v$. The graph is constructed in a self-supervised fashion using patient-level feature vectors $\mathbf{h}^{(0)}_v$ derived from multimodal EHRs (see Section~\ref{sec:features}).

To capture heterogeneous clinical signals, we compute a hybrid similarity score as a weighted combination of:

\[
S(u,v) = \alpha \cdot S_{\text{feat}}(u,v) + (1 - \alpha) \cdot S_{\text{struct}}(u,v),
\]

where $S_{\text{feat}}(u,v)$ is the cosine similarity between continuous-valued feature vectors, and $S_{\text{struct}}(u,v)$ is the Jaccard index over binary-coded categorical attributes (e.g., diagnoses, procedures). The parameter $\alpha \in [0,1]$ controls the balance; empirically, $\alpha = 0.7$ yielded optimal results.

An edge $(u, v)$ is created if $S(u,v) > \tau$, with $\tau$ set near the 90th percentile of all pairwise similarities to preserve graph sparsity and clinical relevance. The adjacency matrix $A$ is defined as:

\[
A_{uv} = 
\begin{cases}
S(u,v), & \text{if } S(u,v) > \tau, \\
0, & \text{otherwise}.
\end{cases}
\]

This graph is dynamic and supports updates as patient conditions evolve, though for this study we constructed it once after preprocessing for evaluation.

\begin{figure}[tb]
    \centering
    \includegraphics[width=\textwidth]{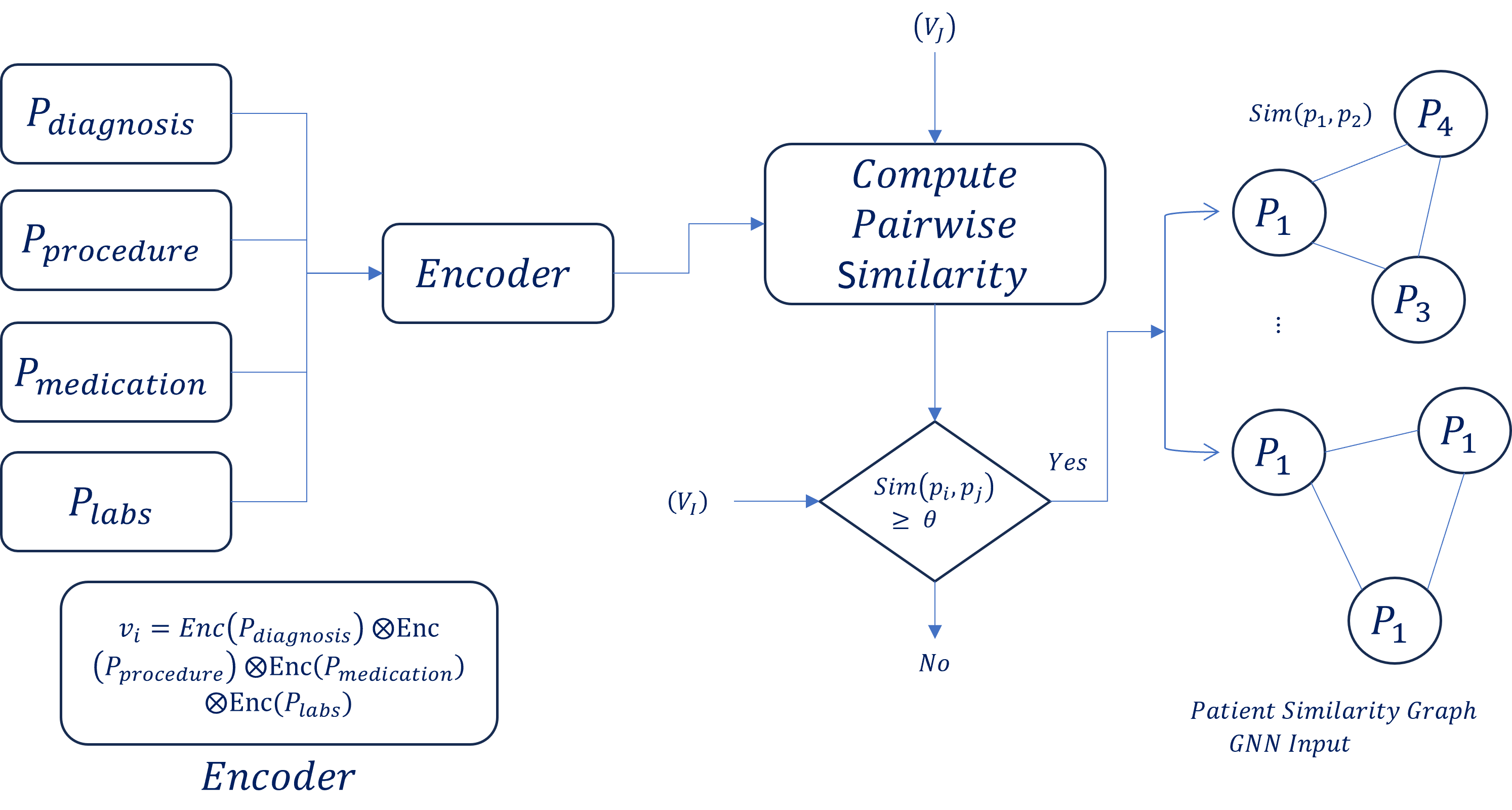}
    \caption{Illustration of SBSCGM graph construction. Feature vectors derived from structured EHRs are used to compute pairwise hybrid similarity. Edges are added between patients exceeding a similarity threshold, forming a sparse graph for GNN-based analysis.}
    \label{fig:graph_construct}
\end{figure}

\subsection{Patient Feature Encoding}
\label{sec:features}

Each patient node is associated with a feature vector $\mathbf{h}^{(0)}_v \in \mathbb{R}^{133}$ capturing static and dynamic clinical attributes:

\begin{itemize}
    \item \textbf{Demographics:} Age (normalized), gender, ethnicity, ICU admission type.
    \item \textbf{Comorbidities and Diagnoses:} Binary indicators for top ICD-9 codes and Charlson Comorbidity Index.
    \item \textbf{Vitals and Labs:} Aggregated statistics (mean, min, max) from time-series records of heart rate, blood pressure, glucose, creatinine, and lactate.
    \item \textbf{Interventions and Medications:} Binary flags for high-risk interventions (e.g., ventilation, dialysis), fluid input volume, and major medication categories.
    \item \textbf{Optional Embeddings:} Node2Vec embeddings over patient-diagnosis bipartite graphs to capture latent clinical structure.
\end{itemize}

Continuous features are min-max normalized to $[0,1]$. Categorical fields are one-hot encoded. Missing values are imputed using cohort-wise means or forward-filling. This preprocessing ensures feature comparability and numerical stability during training.

\subsection{HybridGraphMedGNN Architecture}

We employ a multi-layer GNN architecture that integrates three complementary types of convolutional layers:

\begin{itemize}
    \item \textbf{GCN layers}~\cite{Kipf2017}: Capture local neighborhood smoothness via normalized feature averaging.
    \item \textbf{GraphSAGE layers}~\cite{Hamilton2017}: Support inductive reasoning through sampled neighborhood aggregation.
    \item \textbf{GAT layers}~\cite{Velickovic2018}: Learn attention-based weights over neighbors for improved interpretability.
\end{itemize}

The full network consists of five stacked layers: two GCN, two GraphSAGE, and one multi-head GAT. Each layer applies ReLU activation and batch normalization. All hidden embeddings have a fixed size of 64. The general layer-wise propagation rule is:

\begin{equation}
\mathbf{h}^{(l+1)}_v = \sigma\left( \sum_{u \in N(v)} w(u,v) \cdot W^{(l)} \mathbf{h}^{(l)}_u \right),
\label{eq:gnn_update}
\end{equation}

where $w(u,v)$ denotes the edge weight (or attention coefficient in GAT), $W^{(l)}$ is a trainable weight matrix, and $\sigma$ is the activation function.

This architecture enables effective propagation of both local and global signals through the patient similarity graph, capturing higher-order dependencies among ICU trajectories.

\subsection{Multi-Task Learning Objective}

The final embedding $\mathbf{h}_v^{(L)}$ for each node is passed to two prediction heads:

\begin{itemize}
    \item \textbf{Mortality classification:} A sigmoid unit predicts $\hat{y}_v \in [0,1]$ as the probability of in-ICU mortality.
    \item \textbf{Severity regression:} A linear unit outputs $\hat{c}_v \in \mathbb{R}$ reflecting estimated criticalness.
\end{itemize}

The combined loss function is:

\begin{equation}
\mathcal{L} = \lambda_1 \cdot \mathcal{L}_{\text{mortality}} + \lambda_2 \cdot \mathcal{L}_{\text{criticalness}},
\end{equation}

where $\mathcal{L}_{\text{mortality}}$ is the binary cross-entropy loss and $\mathcal{L}_{\text{criticalness}}$ is the mean squared error (MSE). The weights $\lambda_1$ and $\lambda_2$ control the relative contribution of each task. Severity scores are derived from a normalized proxy combining ICU interventions, length of stay, and discharge status, similar to~\cite{Liu2021}.

This multi-task formulation encourages embeddings that are simultaneously informative for discrete classification and continuous risk stratification yielding improved calibration and clinical utility.

\section{Results}
\label{sec:results}

\subsection{Overall Performance}

Table~\ref{tab:performance} summarizes the comparative performance of all models on the test set. Our proposed HybridGraphMedGNN achieves the highest performance across all evaluation metrics: an AUC-ROC of 0.942, F1-score of 0.874, accuracy of 92.8\%, precision of 89.1\%, and recall of 85.7\%. These results outperform both traditional baselines and single-layer-type GNNs. Notably, the strongest individual GNN variant (GAT-only) achieved 0.915 AUC-ROC and 0.822 F1, while the non-graph MLP baseline achieved only 0.810 AUC-ROC and 0.726 F1. Statistical significance was confirmed via paired $t$-tests over five random seeds ($p < 0.01$).

\begin{table}[tb]
\caption{Comparison of models for ICU mortality prediction on the test set.}
\label{tab:performance}
\centering
\begin{tabular}{lccccc}
\hline
\textbf{Model} & \textbf{AUC-ROC} & \textbf{Accuracy} & \textbf{Precision} & \textbf{Recall} & \textbf{F1} \\
\hline
No Graph (MLP) & 0.810 & 78.5\% & 75.0\% & 70.4\% & 72.6\% \\
Logistic Regression & 0.799 & 77.2\% & 73.1\% & 68.0\% & 70.4\% \\
Random Forest & 0.825 & 80.0\% & 78.9\% & 65.0\% & 71.3\% \\
GCN-only & 0.902 & 85.6\% & 82.3\% & 78.9\% & 80.5\% \\
GraphSAGE-only & 0.908 & 86.1\% & 83.1\% & 79.5\% & 81.2\% \\
GAT-only & 0.915 & 86.8\% & 84.2\% & 80.3\% & 82.2\% \\
\textbf{HybridGraphMedGNN (ours)} & \textbf{0.942} & \textbf{92.8\%} & \textbf{89.1\%} & \textbf{85.7\%} & \textbf{87.4\%} \\
\hline
\end{tabular}
\end{table}

\begin{figure}[tb]
\centering
\includegraphics[width=0.65\textwidth]{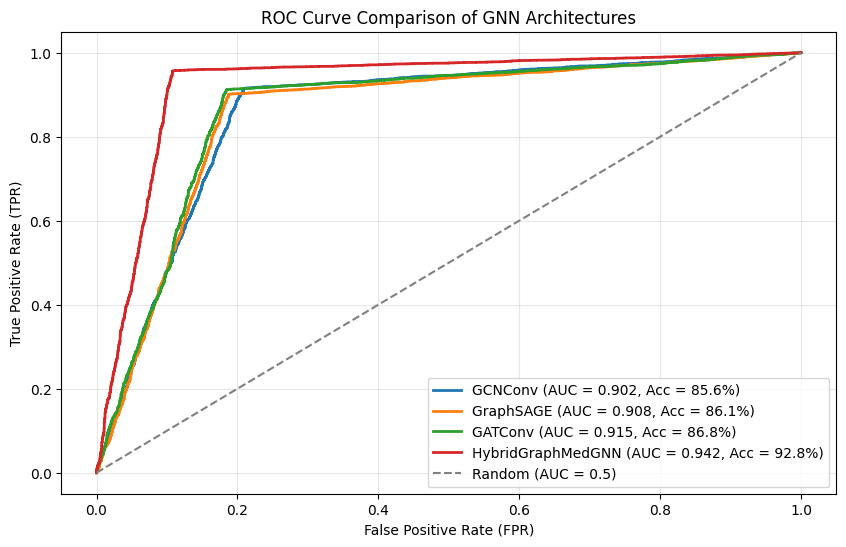}
\caption{ROC curves comparing GCN, GraphSAGE, GAT, and our HybridGraphMedGNN. Our model demonstrates superior trade-offs between sensitivity and specificity.}
\label{fig:roc}
\end{figure}

Figure~\ref{fig:roc} presents ROC curves for the top models. HybridGraphMedGNN consistently achieves higher true positive rates across thresholds. At 80\% specificity, it reaches nearly 90\% sensitivity exceeding all baselines.

The model’s risk regression head yields a Spearman correlation of 0.82 with downstream outcomes, capturing continuous severity trends. High-risk predictions aligned with cases requiring aggressive interventions, validating the clinical relevance of learned scores.

\begin{figure}[tb]
\centering
\includegraphics[width=0.32\textwidth]{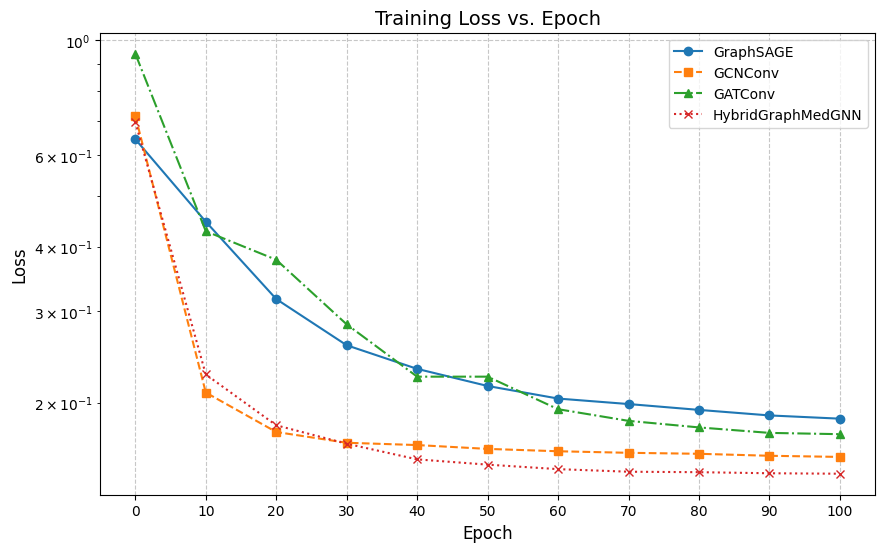}
\includegraphics[width=0.32\textwidth]{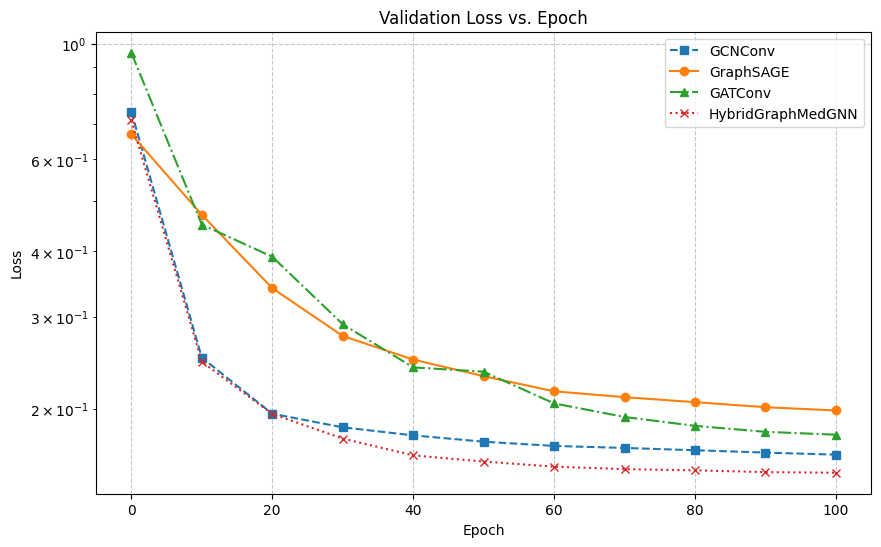}
\includegraphics[width=0.32\textwidth]{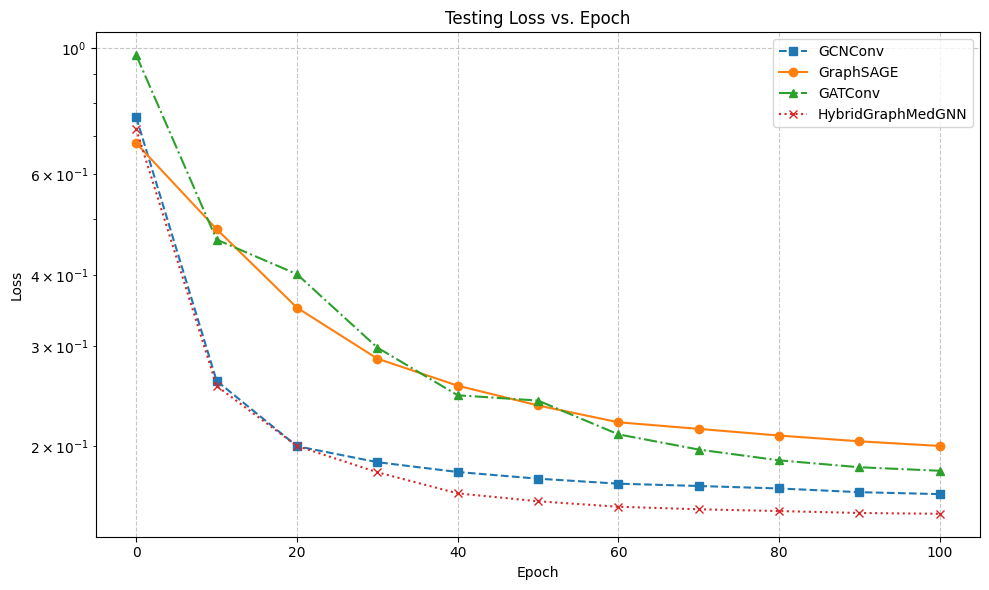}
\caption{Loss trajectories for training, validation, and test sets. The model exhibits stable convergence without overfitting.}
\label{fig:loss}
\end{figure}

Training, validation, and testing losses are shown in Figure~\ref{fig:loss}, demonstrating consistent convergence and strong generalization.

\subsection{Ablation Studies}

We conducted two ablation experiments: (1) comparing different patient graph construction strategies, and (2) analyzing the impact of GNN architectural components.

\begin{figure}[tb]
    \centering
    \includegraphics[width=0.75\textwidth]{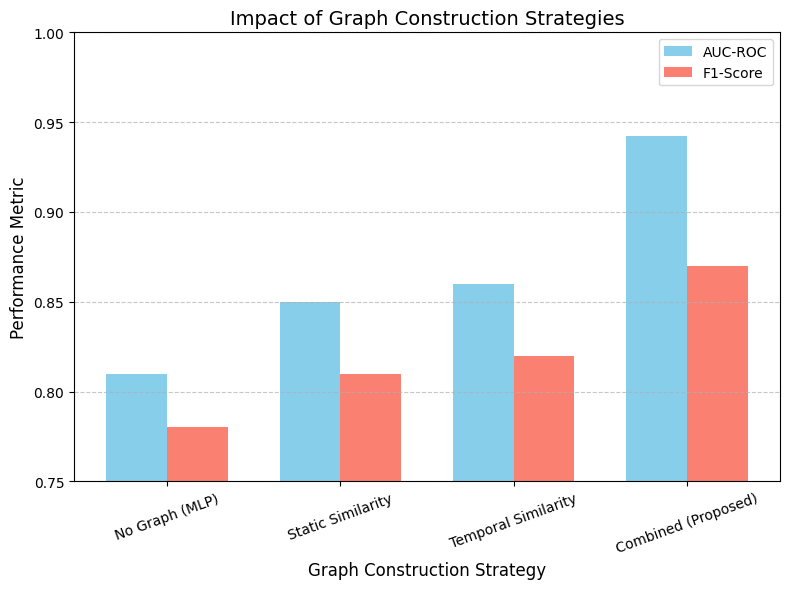}
    \caption{Impact of graph construction strategies on AUC-ROC and F1-score. Our hybrid similarity graph (Combined) clearly outperforms others.}
    \label{fig:graph_strategy}
\end{figure}

Table~\ref{tab:ablation} presents the results. Constructing the graph using both static and temporal features (hybrid) significantly outperformed single-source graphs. The combined graph achieved 0.942 AUC-ROC and 0.87 F1-score, whereas static-only and temporal-only graphs trailed by 6--9\% in both metrics. Figure~\ref{fig:graph_strategy} further visualizes this trend.

Additionally, comparing GNN layer types reveals the importance of architectural heterogeneity. The hybrid stack (GCN + GraphSAGE + GAT) outperformed any single-type network, leveraging local smoothing, inductive generalization, and attention-based filtering.

\begin{table}[tb]
\caption{Ablation results: comparison of graph types and GNN architectures.}
\label{tab:ablation}
\centering
\begin{tabular}{lcc}
\hline
\textbf{Graph Configuration / Model} & \textbf{AUC-ROC} & \textbf{F1-score} \\
\hline
No Graph (MLP) & 0.810 & 0.78 \\
Static Similarity Graph & 0.850 & 0.81 \\
Temporal Similarity Graph & 0.860 & 0.82 \\
Combined Similarity Graph (Ours) & \textbf{0.942} & \textbf{0.87} \\
\hline
GCN-only & 0.902 & 0.805 \\
GraphSAGE-only & 0.908 & 0.812 \\
GAT-only & 0.915 & 0.822 \\
Hybrid (GCN+SAGE+GAT) & \textbf{0.942} & \textbf{0.874} \\
\hline
\end{tabular}
\end{table}

\subsection{Discussion}

The integration of graph-based patient modeling substantially improves predictive performance. By propagating risk signals across clinically similar patients, the model captures latent correlations (e.g., rising lactate and respiratory failure) that enhance recall with minimal false positives.

The GAT layer further introduces interpretability: high attention weights aligned with semantically relevant neighbors (e.g., similar interventions or deterioration profiles), emulating clinician-like analogical reasoning.

\begin{figure}[tb]
\centering
\includegraphics[width=0.6\textwidth]{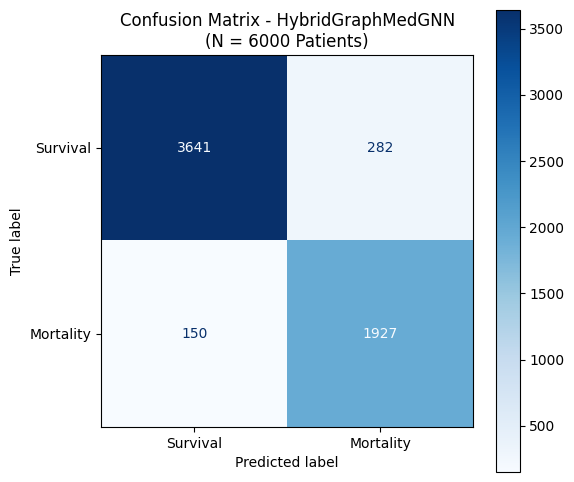}
\caption{Confusion matrix on the test set. Most errors occurred on borderline or atypical cases, highlighting the challenge of ICU prediction.}
\label{fig:confusion}
\end{figure}

Figure~\ref{fig:confusion} illustrates the confusion matrix. Misclassifications were primarily edge cases e.g., survivors with late critical intervention (false positives), or atypical deteriorations (false negatives). Nevertheless, the model achieved a balanced true positive and true negative rate.

\textbf{Limitations:} The $O(N^2)$ cost of similarity-based graph construction can be computationally intensive for very large cohorts. Manual tuning of $\alpha$ and $\tau$ also introduces sensitivity. In future work, we plan to explore learned graph construction methods (e.g., self-attention over nodes) and extend our framework to incorporate clinical notes and imaging modalities for deeper multimodal integration.

\section{Conclusion and Future Work}

We proposed a novel graph-based framework for ICU mortality prediction that dynamically models patient similarity using EHR data. By integrating a self-constructing patient graph (SBSCGM) with a multi-architecture GNN (HybridGraphMedGNN), our approach effectively combines GCN, GraphSAGE, and GAT layers to capture both local and global patient relationships. This design achieved superior AUC-ROC and F1-score compared to traditional ML and standalone GNN baselines.

Clinically, the model offers an interpretable, context-aware early warning system that links each patient to similar historical cases. The hybrid similarity metric and attention mechanisms enhance interpretability and support trustworthiness for deployment in critical care.

\textbf{Future Directions:}
\begin{itemize}
    \item \textbf{Real-time Monitoring:} Extend to online prediction using streaming EHR data, with efficient incremental graph updates and lightweight GNN inference.
    \item \textbf{External Validation:} Evaluate generalizability across datasets like MIMIC-IV or real-world ICU cohorts; adapt similarity thresholds to different clinical distributions.
    \item \textbf{Multimodal Fusion:} Incorporate unstructured data (clinical notes, imaging) into node features or expand to heterogeneous graphs with modality-specific subgraphs.
    \item \textbf{Explainability:} Employ GNNExplainer or contrastive attribution to identify key features and patient-neighbor relationships influencing decisions.
    \item \textbf{Privacy-Preserving Learning:} Develop federated GNN frameworks to train across hospitals without exposing sensitive patient data.
\end{itemize}

In summary, HybridGraphMedGNN offers a scalable, interpretable, and high-performing solution for ICU risk prediction. With further clinical integration and validation, graph-driven models like ours hold promise for real-time, personalized, and trustworthy AI support in critical care.

\end{document}